\pdfoutput=1
\documentclass[11pt]{article}

\usepackage{authblk}

\usepackage{acl}

\usepackage{times}
\usepackage{latexsym}

\usepackage[T1]{fontenc}
\usepackage[utf8]{inputenc}

\usepackage{microtype}

\usepackage{graphicx}
\usepackage{amsmath}
\usepackage{bm}
\usepackage{multirow}
\usepackage{booktabs}
\usepackage{subcaption}
\usepackage{CJKutf8}
\usepackage{caption}
\usepackage{amssymb}

\usepackage{stfloats}
\usepackage{ulem}
\title{InterHT: Knowledge Graph Embeddings by Interaction between Head and Tail Entities}

\author[1,2]{\bf{Baoxin Wang}}
\author[2]{\bf{Qingye Meng}}
\author[2]{\bf{Ziyue Wang}}
\author[2]{\bf{Honghong Zhao}}
\author[2]{\bf{Dayong Wu}}
\author[1]{\\ \bf{Wanxiang Che}}
\author[2,3]{\bf{Shijin Wang}}
\author[2]{\bf{Zhigang Chen}}
\author[2]{\bf{Cong Liu}}
\affil[1]{Research Center for SCIR, Harbin Institute of Technology, Harbin, China}
\affil[2]{State Key Laboratory of Cognitive Intelligence, iFLYTEK Research, China}
\affil[3]{iFLYTEK AI Research (Hebei), Langfang, China}
\affil[ ]{\texttt {\{bxwang2,qymeng5,zywang27,dywu2,sjwang3,zgchen,congliu2\}@iflytek.com}}
\affil[ ]{\texttt {car@ir.hit.edu.cn}}

\begin{document}
\maketitle
\begin{abstract}
Knowledge graph embedding (KGE) models learn the representation of entities and relations in knowledge graphs. Distance-based methods show promising performance on link prediction task, which predicts the result by the distance between two entity representations. However, most of these methods represent the head entity and tail entity separately, which limits the model capacity. We propose two novel distance-based methods named InterHT and InterHT+ that allow the head and tail entities to interact better and get better entity representation. Experimental results show that our proposed method achieves the best results on ogbl-wikikg2 dataset.
\end{abstract}

\section{Introduction}
Knowledge graph embedding (KGE) models learn the representation of entities and relations in the knowledge graphs (KGs). A large number of KGE methods have been proposed to solve the problem of knowledge graph completion. KGE can also be applied to various applications such as recommender systems, question answering and query expansion. 

Nowadays, KGE methods have been widely studied. One of the most popular methods for KGE is the distance-based methods, such as transE \cite{bordes2013translating}, transH \cite{wang2014knowledge}, RotatE \cite{sun2019rotate} and PairRE \cite{chao2020pairre}. These methods represent head and tail entities as deterministic vectors separately. This static entity representation limits the learning capacity of models, which prohibits models to better learn large-scale KG.

In natural language understanding tasks, the contextual representation of words is usually important for the downstream tasks. Inspired by this, we propose a novel distance-based method called InterHT which can represent the head and tail entities by combining the information of the tail and head entities, respectively. Concretely, for the head entity embedding, we element-wise multiply the original head entity representation with an auxiliary tail entity vector to obtain the final representation of the head entity. By this way, we can incorporate the information of the tail entity into the head entity representation. Similarly, we can use the same method to obtain the final representation of the tail entity.

We validate our model on the ogbl-wikikg2 dataset \cite{hu2020open}. The InterHT achieve 67.79\% and the InterHT+ achieve 72.93\% in MRR. The experimental results show that our proposed approach effectively improves the model capacity and achieves the best performance on the dataset.

\section{Related Work}
The link prediction task on KGs is conventionally accomplished via KGE approaches. Generally, these approaches can be categorised into two families, distance-base models and bilinear models. Distance-based models represent the fact \emph{(h, r, t)} in embedding spaces and measure the distance between entities. Usually, they approximate the tail entity vector \textbf{t} via translating the head entity vector \textbf{h} by the relation vector \textbf{r}, and measure the distance between this translated entity and the actual tail entity. The closer distance indicates the higher plausibility that the fact \emph{(h, r, t)} holds. TransX series models, such as TransE/H/R/D \cite{bordes2013translating,wang2014knowledge,lin2015learning,ji2015knowledge}, manipulate in the real space. Among the series, TransE represents entities and relations in a single space for 1to1 mapping. Its extensions employ hyperplane (TransH) and multiple embedding spaces (TansR, TranD, etc) to handle complex relations. RotatE \cite{sun2019rotate} introduces the rotation translation in complex space for modeling more complicated relation patterns. 

To gain better expressiveness, researchers further exploit high dimension spaces, directed relations and multiple relation vectors. For example, GC-OTE \cite{tang2019orthogonal} performs orthogonal transformation and enriches the representation with directed relation vectors, and PairRe \cite{chao2020pairre} uses two vectors to represent a single relation. While entity and relation vectors in these embedding spaces are expressed as deterministic vectors, some methods (TransG \cite{xiao2015transg}, etc.) emphasis the uncertainty of nodes and links via Gaussian distribution. The amount of parameters of the above models grow markedly alone with the improvement of performance, yet they could still suffer from out-of-vocabulary (OOV) problem. NodePiece \cite{galkin2021nodepiece} proposes to tokenise entity nodes into sets of anchors which dramatically reduces the parameter size and provides better generalization to unseen entities.

Instead of measuring the distance, bilinear models scoring the candidates by semantic similarity. RESCAL \cite{10.5555/3104482.3104584}, TATEC \cite{10.5555/3120260.3120289}, DisMult \cite{yang2014embedding} and HolE \cite{10.5555/3016100.3016172} represent entities in real space, and interact head and tail entities according to matrix decomposition to learn the relation matrix. ComplEx \cite{trouillon2016complex} extends the embedding space to complex domain to take advantage of the Hermitian dot product. Since Hermitian product is not commutative, it helps to better capture asymmetric relations. Other approaches measure the similarity directly through deep neural networks and graph neural networks, for instance, SME \cite{10.1007/s10994-013-5363-6}, NAM \cite{DBLP:journals/corr/Liu0LWH16}, GNNs \cite{li2016gated, 10.5555/3305381.3305512, xu2018how}, etc. 

Apart from the commonly used distance-based models and bilinear models, automated machine learning methods, pretrained language models and some retrieval-reranking methods are also used in KG completion tasks. For instance, AutoSF \cite{zhang2020autosf} manages to design and search for better data-dependant scoring functions automatically; KG-BERT \cite{DBLP:journals/corr/abs-1909-03193} treats the fact triples as text sequences and converts KG completion tasks into classification tasks; \newcite{cao-etal-2021-knowledgeable} discuss and compare several prompt-based pre-training methods on KGs; \newcite{lovelace-etal-2021-robust} propose a robust KG completion method with a student re-ranking network.  

\begin{table*}[!tp]
\begin{center}
%\scalebox{0.85}{
\begin{tabular}{lccccccc}
\toprule
Dataset & Task & Nodes & Relations & Edges & Train & Validation & Test \\
\midrule
OGB WikiKG2 & LP & 2,500,604 & 535 & 17,137,181 & 16,109,182 & 429,456 & 598,543 \\
\bottomrule
\end{tabular}%}
\end{center}
\caption{Dataset statistics. LP means link prediction.}\label{dataset}
\end{table*}

\begin{figure*}[t]
\begin{center}
\includegraphics[width=0.9\textwidth]{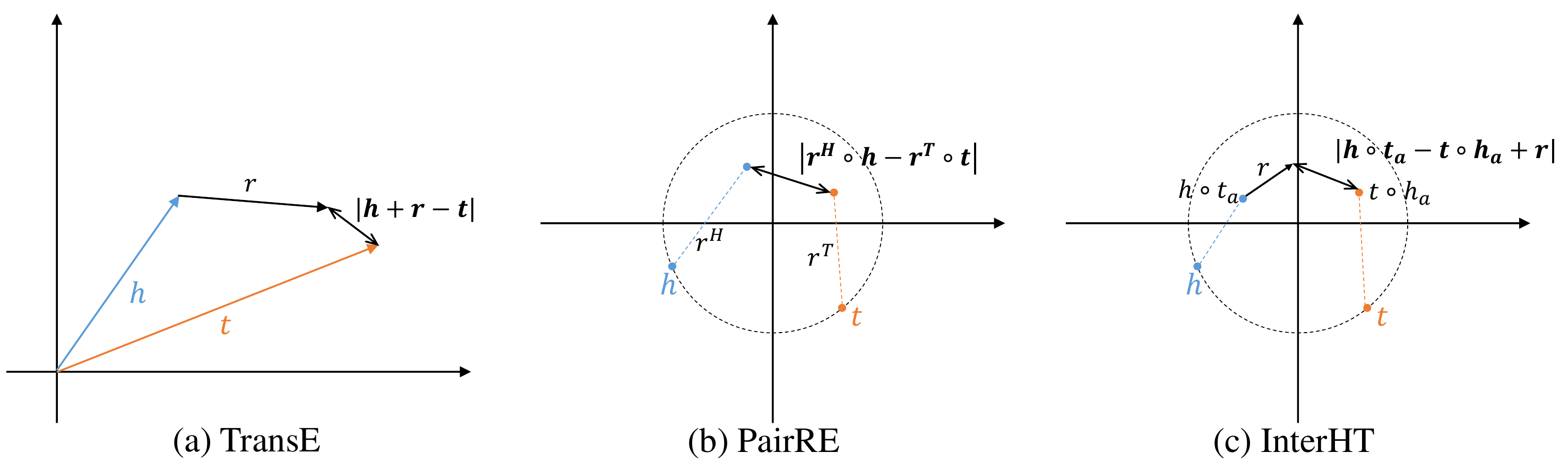}
\caption{Illustration of InterHT.}\label{interht}
\end{center}
\end{figure*}

\section{Method}
In previous research, both head and tail entities are represented independently. We proposed a novel distance-based method named InterHT. The InterHT incorporate the information of the tail entity into the head entity representation, which can improve the model capacity.
\subsection{Entity Representation}
We propose a novel entity representation method named DigPiece (Digraph Piece) which is based on directed coarse-to-fine neighborhood information. First, a subgraph is used to describe a target node, and we use a directed coarse-to-fine way to generate subgraphs. Each subgraph contains four types of nodes: anchors, in-direction neighbors, out-direction neighbors, and center. The anchors and center are the same as StarGraph \cite{li2022stargraph}, but in-direction neighbors and out-direction neighbors are more fine-grained information considering direction. In-direction neighbors refer to a fixed number of nodes sampled from the one-hop head neighborhood of the target node. Out-direction neighbors refer to a fixed number of nodes sampled from the one-hop tail neighborhood of the target node. Then each node of the subgraph is treated as a token, and use a single transformer block to represent the target node. The entity representation is obtained by mean-pooling.

 We use the anchors of NodePiece \cite{galkin2021nodepiece}, but we do not use the BFS method to select the anchors for each node. In order to ensure that the anchors can express the characteristics of the target node better, we will first select all one-hop nodes as anchors. For the two-hop nodes, we count the number of one-hop nodes (excluding anchors) adjacent to each two-hop anchor node. It is preferred to select the two-hop nodes with a larger number.
\subsection{InterHT}
The illustration of the proposed InterHT is shown in Figure \ref{interht}. The final embedding of an entity is generated by interacting over an auxiliary representation of the other entity in the same fact. For the head entity embedding, we element-wise multiply the original head entity representation with an auxiliary tail entity vector to obtain the final representation of the head entity. Similarly, we can use the same method to obtain the final representation of the tail entity. The equations are as follows:

\begin{equation}
\begin{aligned}
||\bf{h} \circ t_{a}-t \circ h_{a}+r||
\end{aligned}
\end{equation}

where $\mathbf{t_a}$ is auxiliary tail entity vector and $\mathbf{h_a}$ is auxiliary tail entity vector. Similar to \citet{long2021triplere}, we add a unit vector $\bf{e}$ to $\bf{t_a}$ and $\bf{h_a}$.

\begin{equation}
\begin{aligned}
||\bf{h} \circ (t_{a} + e)-t \circ (h_{a} + e)+r||
\end{aligned}
\end{equation}

$\bf{h} \circ (t_{a} + e)$ is the representation of the head that combines information of the tail, and $\bf{t} \circ (h_{a} + e)$ is the representation of the tail integrating information of the head. We also use the NodePiece to learn a fixed-size entity vocabulary, which can alleviate the OOV problem.

\subsection{InterHT+}
On the basis of InterHT, we propose InterHT+, which further incorporates the relation embedding into InterHT. The equations are as follows:
\begin{equation}
\begin{aligned}
||&u \cdot \bm{h} \circ \bm{t} + \bm{h} \circ ( u \cdot \bm{r_{h}} + \bm{e}) \\ & - \bm{t} \circ (u \cdot 
\bm{r_t} + \bm{e}) + \bm{r} ||
\end{aligned}
\end{equation}
where $u$ is a constant scalar.

\subsection{Loss Function}
To optimize the model, we utilize the self-adversarial negative sampling loss as the loss function for training \cite{sun2019rotate}. The loss function is as follows:
\begin{equation}
\begin{aligned}
L = & -log\ \sigma(\gamma-d_r(\mathbf{h},\mathbf{t}))\\& -\sum_{i=1}^{n} \frac{1}{k} log\ \sigma (d_r (\mathbf{h_i '}, \mathbf{t_i '})-\gamma )
\end{aligned}
\end{equation}
where $\gamma$ is a fixed margin, $\sigma$ is the sigmoid function, and $(\mathbf{h_i '}, \mathbf{r}, \mathbf{t_i '})$ is the $i$-th negative triplet.

\section{Experiments}
\subsection{Experimental Setup}
\noindent\textbf{Dataset} In this paper we use the ogbl-wikikg2 dataset to validate our approach. ogbl-wikikg2 is extracted from Wikidata knowledge base. One of the main challenges for this dataset is complex relations. The statistics of the dataset are show in Table \ref{dataset}. The MRR metric is adpoted for evaluation.

\begin{table*}[!tp]
\begin{center}
\scalebox{1.}{
\begin{tabular}{lccc}
\toprule
Model & \#Params & Test MRR & Valid MRR \\
\midrule
TransE (500dim) & 1251M & 0.4256 $\pm$ 0.0030 & 0.4272 $\pm$ 0.0030 \\
RotatE (250dim) & 1250M & 0.4332 $\pm$ 0.0025 & 0.4353 $\pm$ 0.0028 \\
PairRE (200dim) & 500M & 0.5208 $\pm$ 0.0027 & 0.5423 $\pm$ 0.0020 \\
AutoSF & 500M & 0.5458 $\pm$ 0.0052 & 0.5510 $\pm$ 0.0063 \\
ComplEx (250dim) & 1251M & 0.5027 $\pm$ 0.0027 & 0.3759 $\pm$ 0.0016 \\
TripleRE & 501M & 0.5794 $\pm$ 0.0020 & 0.6045 $\pm$ 0.0024 \\
\midrule
ComplEx-RP (50dim) & 250M & 0.6392 $\pm$ 0.0045 & 0.6561 $\pm$ 0.0070 \\
AutoSF + NodePiece & 6.9M & 0.5703 $\pm$ 0.0035 & 0.5806 $\pm$ 0.0047 \\
TripleRE + NodePiece & 7.3M & 0.6582 $\pm$ 0.0020 & 0.6616 $\pm$ 0.0018 \\
InterHT + NodePiece & 19.2M & 0.6779 $\pm$ 0.0018 & 0.6893 $\pm$ 0.0015  \\
TripleRE + StarGraph & 86.7M & 0.7201 $\pm$ 0.0011 & 0.7288 $\pm$ 0.0008 \\
InterHT + DigPiece & 156.3M & \textbf{0.7293 $\pm$ 0.0018} & \textbf{0.7391 $\pm$ 0.0023}\\ 
\bottomrule
\end{tabular}}
\end{center}
\caption{Experimental results. The experimental result with DigPiece is based on InterHT+.}\label{experiment_results}
\end{table*}

\noindent\textbf{Baseline Models}
For a sufficient comparison, we employ several typical models for the KG completion tasks and some well-preformed combined methods as baselines, including distance-based models, bilinear models and some enhancing approaches. Baseline models include TransE, RotatE, PairRE, TripleRE, DisMult, ComplEx and AutoSF. Enhancing approaches include NodePiece and Relation prediction (RP)\cite{chen2021relation}. RP refers to the relation type prediction task, which predicts the relation type given the head and tail entities of a fact. We give a brief demonstration of each method here:  

\begin{itemize}
    \item TransE \cite{bordes2013translating} represents the fact \emph{(h, r, t)} as vectors, the head entity vector \textbf{h} is translated by relation vector \textbf{r} and the target is to approximate the tail entity vector \textbf{t}, i.e., $\mathbf{h}+\mathbf{r}\approx \mathbf{t}$. The scoring function is $-\left\| \mathbf{h}+\mathbf{r}-\mathbf{t}\right\|$.
    \item RotatE\cite{sun2019rotate} rotates the head entity vector by the relation vector on a unit circle, i.e., $\mathbf{h}\circ \mathbf{r}=\mathbf{t}$. The scoring function is $-\left\| \mathbf{h}\circ \mathbf{r}-\mathbf{t}\right\|$.
    \item PairRE \cite{chao2020pairre} represent the relation with two vectors, $\mathbf{r}^H$ and $\mathbf{r}^T$, to encode complex relation and multiple relation patterns. The corresponding scoring function is $-\left\| \mathbf{h}\circ \mathbf{r}^H -\mathbf{h}\circ \mathbf{r}^T \right\|$.
    \item TripleRE \cite{long2021triplere} adds a relation translation to the scoring function of PairRE and proposes two versions of the scoring function. We refer to the first version (TripleREv1) for TripleRE baseline, which is as follow, $-\left\| \mathbf{h}\circ \mathbf{r}^h -\mathbf{h}\circ \mathbf{r}^t +\mathbf{r}^m \right\|$. The other is called TripleREv2 and is used in TripleRE+NodePiece baseline, specific, $-\left\| \mathbf{h}\circ (\mathbf{r}^h +u\ast \mathbf{e}) -\mathbf{h}\circ (\mathbf{r}^t +u\ast \mathbf{e}) +\mathbf{r}^m \right\|$, where $u$ is a constant and $\mathbf{e}$ is a unit vector. 
    \item DisMult \cite{yang2014embedding} relation matrix $\mathbf{M}_r$ is restricted to a diagnose matrix and models only the symmetric relations. The scoring function is $\mathbf{h}^\top diag(\mathbf{r})\mathbf{t}$.
    \item ComplEx \cite{trouillon2016complex} represents nodes in complex space and scores the real part of the relation matrix, i.e., $\mathbf{RE}(\mathbf{h}^\top diag(\mathbf{r})\bar{\mathbf{t}})$, where $\bar{\mathbf{t}}$ is the complex conjugate of the tail vector \textbf{t}. 
    \item AutoSF \cite{zhang2020autosf} is a two-stage approach to automatically design a best data-dependent scoring function. It first learns a set of the converged model parameters on the training set, and then search for a better scoring scheme on the validation set.
    \item AutoSF+NodePiece embeds entities with respect to the anchor-based approach and uses AutoSF to design the scoring function.
    \item TripleRE+NodePiece used NodePiece entity embeddings and TripleREv2 scoring function
    \item TripleRE+StarGraph used StarGraph entity embeddings and TripleREv2 scoring function
    \item ComplEx-RP adds the relation prediction probability, $\lambda log P_\theta (r|h,t)$, to ComplEx's training objective.
\end{itemize}

\noindent\textbf{Model Hyperparameters} 
We utilize Adam optimizer with a learning rate of 5e-4. The training batch size is set to 512. We set the maximum number of training steps to 500,000 and validate every 20,000 steps. The dimension of entity embedding of InterHT is set to 200, and the dimension of InterHT+ is set to 512. The negative sample size is 128. The number of anchors for NodePiece is 20,000 and the hyperparameter $u$ for InterHT+ is 0.05.

\subsection{Experimental Results}
The experimental results are shown in Table \ref{experiment_results}. From the experimental results, we can see that InterHT achieves better performance than transE, which indicates that InterHT has a stronger model capacity. After adding NodePiece, InterHT achieves 67.79\% in MRR, which yields a 1.97\% increment over TripleRE. The InterHT+ with DigPiece achieves 72.93\% in MRR. The experimental results show that our proposed approach effectively improves the model capacity and achieves the best results on ogbl-wikikg2 dataset.

\subsection{Conclusion}
In this paper, we propose a novel distance-based knowledge graph embedding model named InterHT. Our proposed method enhance the learning ability of the model with additional head and tail interactions. Experimental results show that our method achieves the best results on ogbl-wikikg2 dataset, reaching 72.93\% in MRR on the test set.

\bibliography{anthology,custom}
\bibliographystyle{acl_natbib}

\end{document}